**Scalable Traffic Signal Controls using Fog-Cloud Based Multiagent Reinforcement Learning**

**Paul (Young Joun) Ha**
Graduate Research Assistant, Center for Connected and Automated Transportation (CCAT), and Lyles School of Civil Engineering, Purdue University, West Lafayette, IN, 47907.
Email: ha55@purdue.edu
ORCID #: 0000-0002-8511-8010

**Sikai Chen\***
Visiting Assistant Professor, Center for Connected and Automated Transportation (CCAT), and Lyles School of Civil Engineering, Purdue University, West Lafayette, IN, 47907.
Email: chen1670@purdue.edu; and
Visiting Research Fellow, Robotics Institute, School of Computer Science, Carnegie Mellon University, Pittsburgh, PA, 15213.
Email: sikaichen@cmu.edu
ORCID #: 0000-0002-5931-5619
(Corresponding author)

**Runjia Du**
Graduate Research Assistant, Center for Connected and Automated Transportation (CCAT), and Lyles School of Civil Engineering, Purdue University, West Lafayette, IN, 47907.
Email: du187@purdue.edu
ORCID #: 0000-0001-8403-4715

**Samuel Labi**
Professor, Center for Connected and Automated Transportation (CCAT), and Lyles School of Civil Engineering, Purdue University, West Lafayette, IN, 47907.
Email: labi@purdue.edu
ORCID #: 0000-0001-9830-2071

## ABSTRACT

Optimizing traffic signal control (TSC) at intersections continues to pose a challenging problem, particularly for large-scale traffic networks. It has been shown in past research that it is feasible to optimize the operations of individual TSC systems or a small number of such systems. However, it has been computationally difficult to scale these solution approaches to large networks partly due to the curse of dimensionality that is encountered as the number of intersections increases. Fortunately, recent studies have recognized the potential of exploiting advancements in deep and reinforcement learning to address this problem, and some preliminary successes have been achieved in this regard. However, facilitating such intelligent solution approaches may require large amounts of infrastructural investments such as roadside units (RSUs) and drones in order to ensure thorough connectivity across all intersections in large networks, an investment that may be burdensome for agencies to undertake. As such, this study builds on recent work to present a scalable TSC model that may reduce the number of required enabling infrastructure. This is achieved using graph attention networks (GATs) to serve as the neural network for deep reinforcement learning, which aids in maintaining the graph topology of the traffic network while disregarding any irrelevant or unnecessary information. A case study is carried out to demonstrate the effectiveness of the proposed model, and the results show much promise. The overall research outcome suggests that by decomposing large networks using fog-nodes, the proposed fog-based graphic RL (FG-RL) model can be easily applied to scale into larger traffic networks.

**INTRODUCTION**
With growing global populations, increased urbanization, and trends of growing automobile ownership, urban transportation networks are increasingly subjected to traffic congestion. The consequential loss of time, increased emissions, and reduced safety in urban transportation can be expected to grow along with increased congestion. The optimization and control of traffic signals represent a key strategy for the management of traffic congestion and improving traffic conditions in urban areas. According to the Federal Highway Administration (FHWA), poor signal timing can account for up to 10% of traffic congestion (*1*). Further, implementation of advanced traffic signal control (TSC) systems in Phoenix, Arizona has seen reductions in traffic collisions by 6.7%, travel times by 11.4%, and delay by 24.9% (2). Therefore, developing and deploying advanced TSC systems can be integral to improving urban traffic conditions.

Traffic signal control is a domain that has seen much attention and research due to its direct impact on social and commercial activities. Broadly, TSC can be classified into two categories: fixed-time traffic control and real-time traffic control. Fixed-time traffic control typically uses a pretimed program that controls the cycle and split times. Webster (1958) was one of the earliest researchers to present a fixed-time control model, which aimed to minimize average delay of vehicles (3). For traffic flow conditions that are stable and do not exhibit randomness, fixed-time traffic control is well-suited. However, for traffic flow conditions that exhibit high levels of stochasticity and instability, fixed-time traffic control models are unsuitable due to their static nature. A better alternative are real-time traffic control models that are responsive to traffic conditions. Traffic control strategies can be made using real-time traffic data, allowing signals to adjust accordingly with unstable and/or stochastic traffic flow. A widely used real-time traffic controller is the actuated signal, which regulate its cycle and timings according to the detector and sensor inputs of the real-time traffic. While many applications of actuated signals have been developed and deployed to great effect, they suffer from the inability to cooperate with many other intersections and do not utilize queues of other phases. Therefore, actuated traffic controllers are unsuitable for addressing network-wide control of urban intersections.

In most urban areas, travel patterns are highly dynamic, and traffic signals are deeply interconnected. Poorly designed signal timings can paradoxically exacerbate congestion, especially when a locally optimal solution is scaled up to large networks. While several studies have shown that traffic signal control (TSC) methods such as actuated and pretimed controls are adequate for small networks (*4, 5*), they cannot be integrated effectively into large networks. With the imminent emergence of robust vehicular connectivity and automation technologies, many solutions on traffic signal control leveraging such technologies are being studied. Guo et al., (2019) presented six types of connected and automated vehicle- (CAV) based traffic control methods including an improved actuated system that utilizes CAV data (6). However, the question of when the CAVs will be deployed into the real-world is still largely debated. As such, this study aims to provide an intelligent, scalable traffic control model that can be integrated into large, urban networks without utilizing CAVs directly.

The prospect of scaling small-intersection TSC solutions to larger networks has been a persistent challenge that has been addressed using a variety of optimization algorithms. In recent years, there has been pronounced interest in the investigation of other solutions methods, and this new direction is motivated by advancements not only in computer hardware and software, including computing power, but also in techniques and technologies for data management and analytics including artificial intelligence and machine learning. For example, multi-threaded, multi-core central processing units (CPUs) such as the Ryzen Threadripper series with up to 64-cores and 128-threads have become more widely available for consumer use. Combined with advances in graphics processing units (GPUs) and large video random access memory (vRAM) capabilities, training deep reinforcement learning models has become much more efficient in recent years. It is acknowledged that deep learning and reinforcement learning concepts were introduced several decades ago (7, *8*). However, recent advancements in computational capabilities have made their application more feasible and therefore have fostered a new generation of deep and reinforcement learning algorithms in continuous and discrete control (*9*, *10*). Alongside emerging smart infrastructural technologies that facilitate real-time data collection and sharing such as road-side units (RSUs) and drones, the implementation and deployment of scalable TSCs and other intelligent transportation systems have become increasingly feasible.

For these reasons, deep reinforcement learning (DRL) based approaches to solving TSC problems in large networks has become an increasingly studied topic. Wiering's study was one of the earliest to propose the use of reinforcement learning algorithms for traffic signal control to minimize city-wide congestion (*11*). Prashanth and Bhatnagar proposed reinforcement learning with function approximation for traffic signal control, using Q-learning for adaptive signal control (*12*). Chu et al proposed a multiagent deep reinforcement learning algorithm that could be applied to large-scale networks; they applied an actor critic network to recurrent neural network with long-short term memory (LSTM) (*13*). Wang et al, proposed the cooperative double Q-learning (Co-DQL) model that leverages mean field approximation of all other agents in the network to significantly reduce model complexity and the curse of dimensionality (*4*).

While the aforementioned studies utilize the state-of-the-art DRL approaches for TSC problems, an oft overlooked topic is the resource constraints that may restrict transportation agencies and other government entities from deploying data-facilitating infrastructure such as RSUs and drones. As such, this study presents an alternative perspective to scalable TSC models that can reduce the number of deployed data-facilitating infrastructure. In essence, the proposed model utilizes a graph attention network (GAT) to preserve the topology of the traffic network while focusing on relevant inputs to make decisions. Doing so allows the model to address large networks as well as variable sized inputs. RSUs are deployed in an urban grid-like network, each serving as fog-nodes that collect data via detectors and share with other fog-nodes in its range, utilizing the information to control the phase and duration of the traffic lights in its control. The Q-network utilizes double estimators to approximate $\max_{a} E\{Q_t(s_{t+1}, a)\}$ instead of maximizing over the estimated action values in the corresponding state to approximate the value of the next state (as is the case in standard Q-learning), performance overestimation is avoided. Overall, the model extracts node embeddings from fog node features while also constructing an adjacency matrix that maps the topology of the connected fog nodes, which are passed through the attention layer to be used for the Q-network. To the best of the authors' knowledge, this is the first study that considers preservation of network topology in TSC problems through the use of GATs.

## BACKGROUND FOR THE METHODOLOGY

### Reinforcement Learning

In general, reinforcement learning (RL) utilizes feedback of decisions, observations, and rewards. Deep reinforcement learning (DRL) combines RL with deep learning, which allows for end-to-end training of multilayer models that can solve complex problems. This is particularly useful for sequential decision making such as in robotics, video games, and traffic operations (*10, 14-17*).

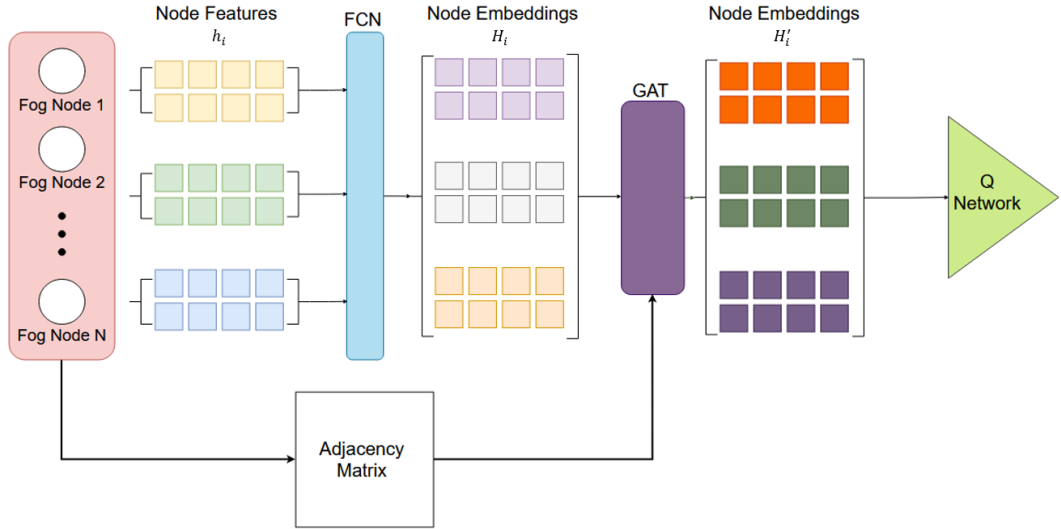

Figure 1: GAT Model Architecture

One of the most popular single-agent RL method is Q-learning. Q-learning is a model-free reinforcement learning approach that can be considered as asynchronous dynamic programming, where agents learn optimal policies in Markovian domains through solving sequential decision-making problems (18). This is achieved through estimating the optimal value, $Q^*(s, a) = \max_{\pi} Q^\pi(s, a)$, for each action $a$ during state $s$. Because most problems have large state and action spaces to learn all action values separately, a parametrized value function $Q(s, a; \theta_t)$ can be learned instead. Thus, the standard Q-learning update for the parameter from taking action $a_t$ in state $s_t$ with observed reward $r_{t+1}$ and the subsequently resulting state $s_t$ is

$$\theta_{t+1} = \theta_t + \alpha \left( Y_t^Q - Q(s_t, a_t; \theta_t) \right) \nabla_{\theta_t} Q(s_t, a_t; \theta_t)$$

where $\alpha$ is the learning rate, and the target $Y_t^Q$ is defined as

$$Y_t^Q \equiv r_{t+1} + \gamma \max_a Q(s_{t+1}, a; \theta_t)$$

where the constant $\gamma \in [0,1)$ is the discount factor adjusting the weight between immediate and later rewards.

Q-learning in multiagent reinforcement learning (MARL) differs primarily in that MARL is based on Markov game instead of a Markov decision process (MDP) (19). Similarly to MDPs, Markov games can be represented as a tuple $(M, \boldsymbol{S}, \boldsymbol{A}_{1,2,\dots,M}, r_{1,2,\dots,M}, p)$, where M is the number of agents, $\boldsymbol{S} = \{s_1, s_2, \dots, s_m\}$ is the set of system states, $\boldsymbol{A}_m$ is the action set of agent $m \in \{1,2,\dots,M\}$, $r_m: \boldsymbol{S} \times \boldsymbol{A_1} \times \dots \times \boldsymbol{A_M} \times \boldsymbol{S} \to \mathbb{R}$ is the reward function for agent $m$, and $p: \boldsymbol{S} \times \boldsymbol{A_1} \times \dots \times \boldsymbol{A_N} \to \mu(\boldsymbol{S})$ is the transition function for moving from one state $s$ to another state $s'$ given action $a_{1,2,\dots M}$. Partially observable Markov games additionally require $\Omega$, the set of observations of the hidden states, and $\mathcal{O}: \boldsymbol{S} \times \Omega \rightarrow \mathbb{R}_{\geq 0}$, the observation probability distribution.

In MARL, each agent learns to choose its actions according to their respective strategies. At each time step, the system state transfer occurs by taking the joint action $a = (a_1, \ldots, a_M)$ under the joint strategy $\pi \triangleq (\pi_1, \ldots, \pi_M)$, and each agent receives their immediate reward from the joint action. For each agent $m$ under joint policy $\pi$ and initial state $s(0) = s \in \boldsymbol{S}$, the expected discounted reward is:

$$V_m^\pi(s) = E^\pi \left\{ \sum_{t=0}^{\infty} \gamma^t r_m(t+1) | s(0) = s \right\}$$

Additionally, the agent-specific average reward can be found as:

$$J_m^\pi(s) = \lim_{T \to \infty} \frac{1}{T} E^\pi \left\{ \sum_{t=0}^{T} r_m(t+1) | s(0) = s \right\}$$

**Graph Neural Networks**

Graph neural networks are able to preserve acyclic and nonacyclic graph topology, which can enhance road network representation particularly in the context of scalable network traffic signal control ($21 - 23$). Deep reinforcement learning requires a strong neural network architecture for forward and backpropagation for model training (24). Graph convolutional networks (GCNs) can serve as powerful neural networks that can address graph data for deep reinforcement learning ($25$, $26$). The nodes of a GCN layer aggregates its own observed states and those of its neighbors into embeddings. Given different relational graphs, the message propagation is as follows ($26$):

$$h_i^{l+1} = \varsigma \left( \Sigma_{m \in \mathcal{M}_i} g_m \left( h_i^l, h_j^l \right) \right)$$

where $h_i^l \in \mathbb{R}^{d^{(l)}}$ denotes the hidden state of node $v_i$ in the $l^{th}$ layer of the neural network, $d^{(l)}$ is layer dimensions, $\mathcal{M}_i$ is the set of incoming messages, and $g_m(\cdot)$ is the transformation for the message from the nodes.

In essence, these node embeddings can address problems caused by variable length inputs to perform various sequential learning tasks given graph data, and error terms can be used to backpropagate to perform the requisite gradient descent for parameter tuning purposes.

**METHODOLOGY**

**DRL Model Architecture**

The fog-based graphic RL (FG-RL) model for TSC presented in this paper employs a scalable and decentralized methodology. The graphical structure of the network topology is preserved with traffic signals and intersections, along with their relative adjacencies. The fog arrangement determines the topology of the connected entities and the number of the connected intersections within its range. Therefore, the adjacency matrix containing the relative adjacencies and connectivity of intersections vary corresponding to how the RSUs (and in turn, the fog nodes) are dispersed in the network and how many intersections each RSU oversees. In DRL architecture, each RSU is represented as a fog node, which serves as an agent that makes decisions to select traffic signal phases for each of the intersections it oversees, with an overall goal to reduce congestion.

The network topology and information attention are modeled using GAT. The fog node can oversee multiple intersections, some of which may have few or no queued vehicles. Therefore, it must learn to divert attention away from relatively uncongested intersections and focus more on congested intersections. However, a given intersection's congestion levels can vary drastically between episodes or even across different time-steps in one episode. As a result, applying an attention model can facilitate the learning process under conditions when such variations exist.

Each fog node $i$ produces node embeddings that encode node features $h_i$. The state is a tuple of $N \times F$ node feature matrix $X_t$ and an $N \times N$ adjacency matrix $A_t$, where $N$ is the total number of nodes, and $F$ is the number of features in each node. The feature matrix considers the states consistent with those in the literature (*7, 14*), namely, (i) the cumulative delay of the first vehicle in each incoming lane at an intersection, and (ii) the total number of approaching vehicles in each incoming lane.

The network architecture is shown in Figure 1. At each time-step $t$, the node feature matrix $X_t$ is fed as the input into a fully connected encoder denoted $\varphi$ that generates node embeddings $H_t$ in $d$ dimensional embedding space $\mathcal{H} \in \mathbb{R}^{N \times d}$

$$H_t = \varphi(X_t) \in \mathcal{H}$$

The node embeddings then are passed through the graph convolution with attention mechanism.

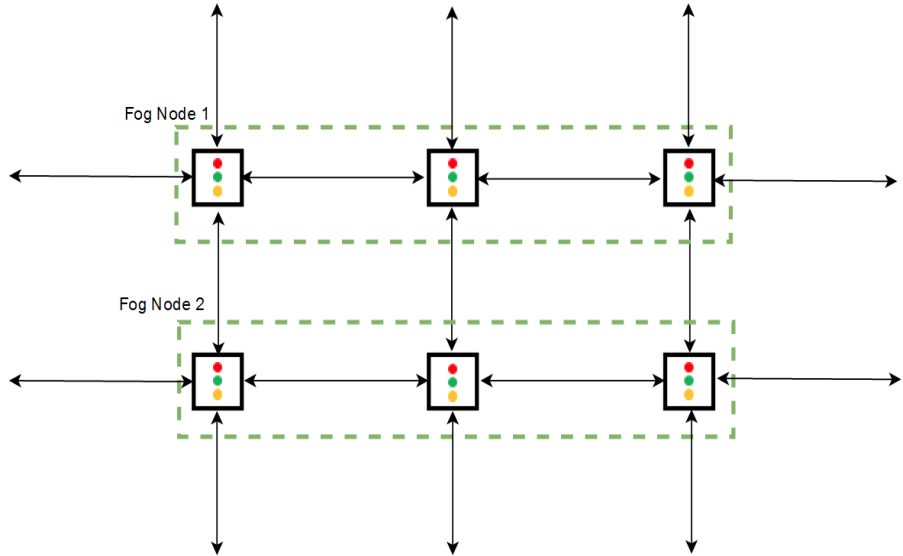

Figure 2: Small TSC Network

The adjacency matrix is weighted using the attention mechanism:

$$H_t' = GAT(H_t, A_t) = \alpha H_t W + b$$

where $\alpha_{ij}$ are coefficients computed by the attention mechanism defined in the literature (*28*):

$$\alpha_{ij} = \frac{\exp\left(LeakyReLU\left(\boldsymbol{a}^T[\boldsymbol{W}h_i||\boldsymbol{W}h_j]\right)\right)}{\Sigma_{k\in(\mathcal{N}_i)}\exp\left(LeakyReLU(a^T[\boldsymbol{W}h_i||\boldsymbol{W}h_k])\right)}$$

The output of the GAT layer is then used as inputs to the Q network to obtain the Q values. Further, experience relay and soft target update are utilized to enhance learning (*11, 29*), and the model is trained on randomly sampled batches from a replay buffer. Thus, the architecture can be summarized as follows:

- FCN Encoder $\varphi$: Dense (32) + Dense (32)

- GAT Layer $GAT$: GATConv (32)

- Q Network: Dense (32) + Dense (32) + Dense (64) + Dense (32)

- Output Layer: Dense (5)

## CASE STUDY

The case study utilized the Simulation of Urban MObility (SUMO) for traffic simulation (*30*), an open-source simulator that enables detailed tracking of vehicle and traffic light parameters. For an initial proof of concept, a small 6-node network is considered (Figure 2).

**Network Descriptions**
A small grid network was used for numerical experimentation, as shown in Figure 2. The first setting utilizes a "smart cities" approach, where each intersection is connected via a central controller in a cloud environment. This setting is a fully observable MDP. It must be noted that this is an ideal setting that has no constraints, meaning that all entities are assumed to be connected. While this can be achieved easily in simulation, it will need many connectivity facilitating infrastructure units to ensure that the entire network is connected. Especially in large networks, this can be problematic.

The second setting utilizes the proposed fog-node approach, where intersections are grouped together by a small number of connectivity facilitating infrastructure such as RSUs or drones. Specifically, for this numerical example, two fog nodes are deployed such that the upper horizontal intersections are connected, and the lower horizontal intersections are connected. As previously states, the two benefits of segmenting the whole network into smaller fog nodes is the improved scalability and the possibility of reducing the number of RSUs/drones required to facilitate the intelligent TSC models.

Each westbound and eastbound road segment entering signalized intersections is a two-lane arterial comprised of a through-lane and a left-turn lane. Each northbound and southbound road segment consists of a single through-lane. Vehicles enter each outer road segments (10 total) at a flow rate of 2200 vehicles/hour. The vehicle origins and flows are randomly distributed.

**MDP Settings**
*Action space*
Each fog node controls the three traffic signals in its range. As shown in Figure 2, Node 1 controls the top three signals, and Node 2 controls the bottom three signals. Each signal can take one of five pre-determined phases, as is consistent with most literature and the practice (*7, 14*): east-west straight, east-west left-turn, three straight and left-turn phases for east, west, and north-south.

*State space:*
The local state observed within each fog node is defined as follows:
$$s_{k,t} = \{wait_{k,t}[lane], wave_{k,t}[lane]\}$$
As stated previously, $wait_{k,t}[lane]$ denotes the cumulative delay of the first vehicle for a given lane in an intersection, and $wave_{k,t}[lane]$ denotes the total number of approaching vehicles along each incoming lane.

*Rewards*
The reward function consists of two main penalties:
$$r_1 = wait_{k,t}[lane]$$
$$r_2 = wave_{k,t}[lane]$$

The total reward is the negative weighted sum of the two penalties,
$$r = -\Sigma_{lane}(\sigma_1 r_1 + \sigma_2 r_2)$$
where $\sigma_1, \sigma_2$ are used to scale the two penalties.
This numerical example used $\sigma_1 = 1$ and $\sigma_2 = 0.30$.

**Preliminary Results**
Figure 3 presents a comparison of training results using 2 fog nodes versus a fully observable system.

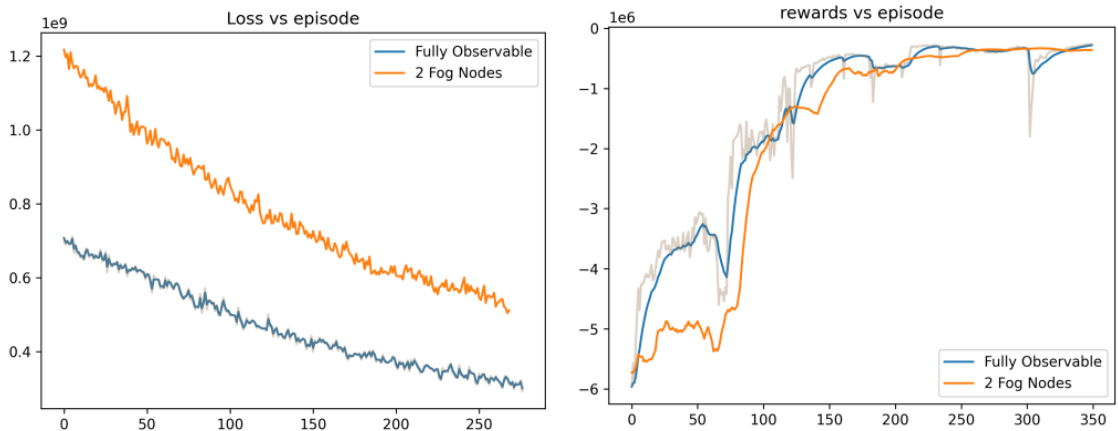

Figure 3: Comparison of Training Results using 2 Fog Nodes vs Fully Observable System

For each setting, the model was trained using a soft target update set at $1e^{-3}$ and a learning rate of $1e^{-5}$. Each model was trained for a total of 100,000 time-steps, with 20,000 time-steps being used for warm-up. Given these training parameters, the training results for the fully observable "smart cities" setting and the fog-node setting are shown in Figure 3. It can be seen that despite the lack of information sharing between the upper and lower intersections, the fog-node setting still performs comparably to a fully observable setting.

However, despite similar training performances, the use of fog nodes results in higher average intersection delay, as shown in Figure 4. Over a 1,000 time-step policy replay, the fully observable model ends with about 150-second average intersection delay. On the other hand, the use of two fog nodes with no communication between them results in almost 300-seconds of average intersection delay at the end of the policy replay.

The primary shortcoming of a fully observable model for traffic signal control problems is that they cannot scale well due to the curse of dimensionality as the number of connected nodes increases. These preliminary results indicate that the use of two separately controlled fog nodes allows for comparable training performance while being more scalable, but at the cost of some performance.

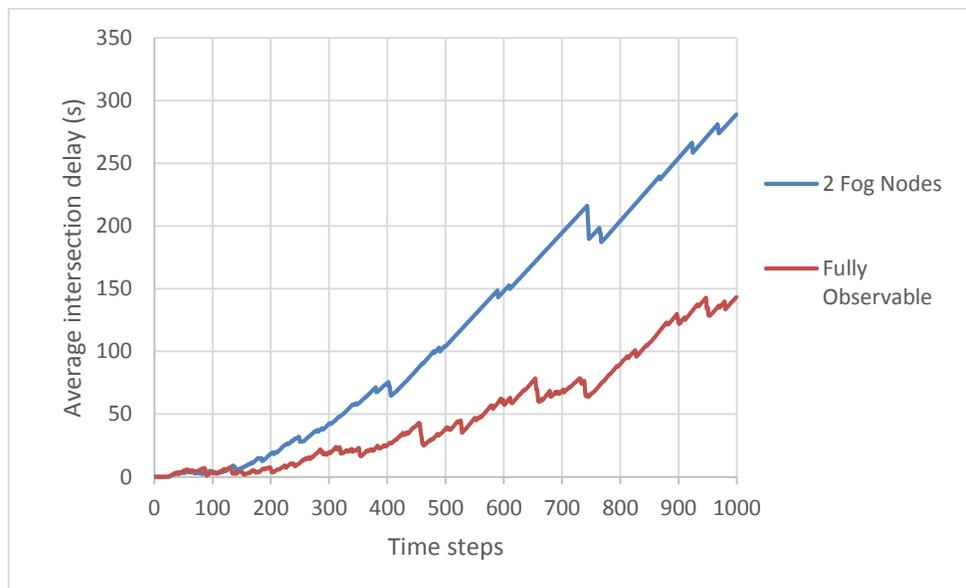

Figure 4: Average intersection delay

## CONCLUSION

In order to create a more easily scalable, intelligent traffic signal control (TSC) model that can be applied to large networks, this paper proposed the use of graph attention networks (GATs) and fog-node architecture. The added benefit of segmenting large networks into smaller fog-nodes includes the possibility of reducing the number of smart infrastructure units required to facilitate the intelligent TSC models. Multiagent reinforcement learning based models for TSC typically can be affected by the curse of dimensionality. The proposed model addresses scalability in two ways: (i) graph attention that only utilizes relevant node features and neighbor node features to reduce the input complexity, and (ii) fog-nodes that break up the large network into manageable sizes (*31*). Preliminary findings show that the proposed model shows promising results that can be scaled into larger networks.

However, their performance in reducing average intersection delay may be relatively inferior compared to a fully observable model. As such, ongoing work on various fog node deployment arrangements and their performance, are expected to provide additional insights on the tradeoff between scalability and performance using the proposed GAT and fog-node architecture. Another promising research direction is to create a simplified or averaged performance within each fog-node to reduce the data size and complexity, thereby allowing fog-nodes to exchange data between each other to make decisions based on other fogs' performances.

## ACKNOWLEDGMENT

This work was supported by Purdue University's Center for Connected and Automated Transportation (CCAT), a part of the larger CCAT consortium, a USDOT Region 5 University Transportation Center funded by the U.S. Department of Transportation, Award #69A3551747105. The contents of this paper reflect the views of the authors, who are responsible for the facts and the accuracy of the data presented herein, and do not necessarily reflect the official views or policies of the sponsoring organization.
This manuscript is submitted for PRESENTATION ONLY at the TRB 2022 Annual Meeting

## AUTHOR CONTRIBUTIONS

The authors confirm contribution to the paper as follows: all authors contributed to all sections. All authors reviewed the results and approved the final version of the manuscript.